\documentclass[letterpaper]{article} 
\usepackage[preprint]{aaai2027} 
\usepackage[hyphens]{url} 
\usepackage{graphicx} 
\usepackage{natbib} 
\usepackage{caption} 
\usepackage{booktabs}
\usepackage{algorithm}
\usepackage{algorithmic}
\usepackage{amsmath}
\usepackage{amssymb}
\usepackage{subcaption}
\usepackage{xcolor}

\title{Staying True to the Origin: Continuous Image Stylization with Smooth Transitions}
\author{
  Rui Xu,
  Hanmo Zhang,
  Songhua Liu\corresponding
}
\affiliations{
  Shanghai Jiao Tong University\\
  chiaroair@gmail.com, hammershock@163.com, liusonghua@sjtu.edu.cn
}

\begin{document}

\maketitle

\begin{abstract}
  Recent advances in generative models have achieved remarkable performance in text- and image-conditioned editing. However, preserving the content of a given image while referencing style patterns from another remains challenging, often leading to uncontrollable stylization results.
  In this paper, we approach image stylization from the perspective of continuous control, aiming to enable modern Diffusion Transformer (DiT)-based multi-reference editing models to (1) faithfully preserve the semantic structure of the content image, (2) render strong stylization effects, and (3) smoothly transition between the two.
  To this end, we propose a simple yet effective two-stage training strategy along with a style-strength-aware spline formulation.
  Specifically, in the first stage, the model is trained to produce strongly stylized outputs while preserving the content semantics as much as possible. In the second stage, with the base model frozen, we learn a set of anchor projectors that map various stylization strengths into the model parameter space. During inference, by performing style-strength-aware spline interpolation in a low-rank space, our method enables continuous control over stylization strength, even though the model is trained with only a few discrete strength levels.
  Extensive experiments demonstrate that our method supports precise and continuous manipulation of stylization strength while generating high-fidelity results with modern DiT models.
  Project page: \url{https://reychiaro.github.io/StyleController}.
\end{abstract}


\section{Introduction}
\label{sec:introduction}

Recent advancements in generative models~\cite{ddpm, flowmatching} have demonstrated profound capabilities in local detail editing~\cite{qwenimage, flux, gpt, gemini, sd3}, particularly when guided by textual and visual modalities via Classifier-Free Guidance (CFG)~\cite{cfg}. Despite their ability to produce high-fidelity images, these models often struggle with global understanding tasks such as style transfer~\cite{vgg}. Fundamentally, style transfer requires generating stylized outputs that simultaneously preserve the semantic structure of a content image and incorporate intricate patterns from a style reference. To address this, several prominent studies~\cite{omnistyle, omniconsistency, ominicontrol} have focused on unleashing the inherent stylization potential of image editing models, employing efficient parameter-tuning techniques like LoRA~\cite{lora} to balance content integrity with stylistic fidelity.

\begin{figure}[t]
  \centering
    \includegraphics[width=\linewidth]{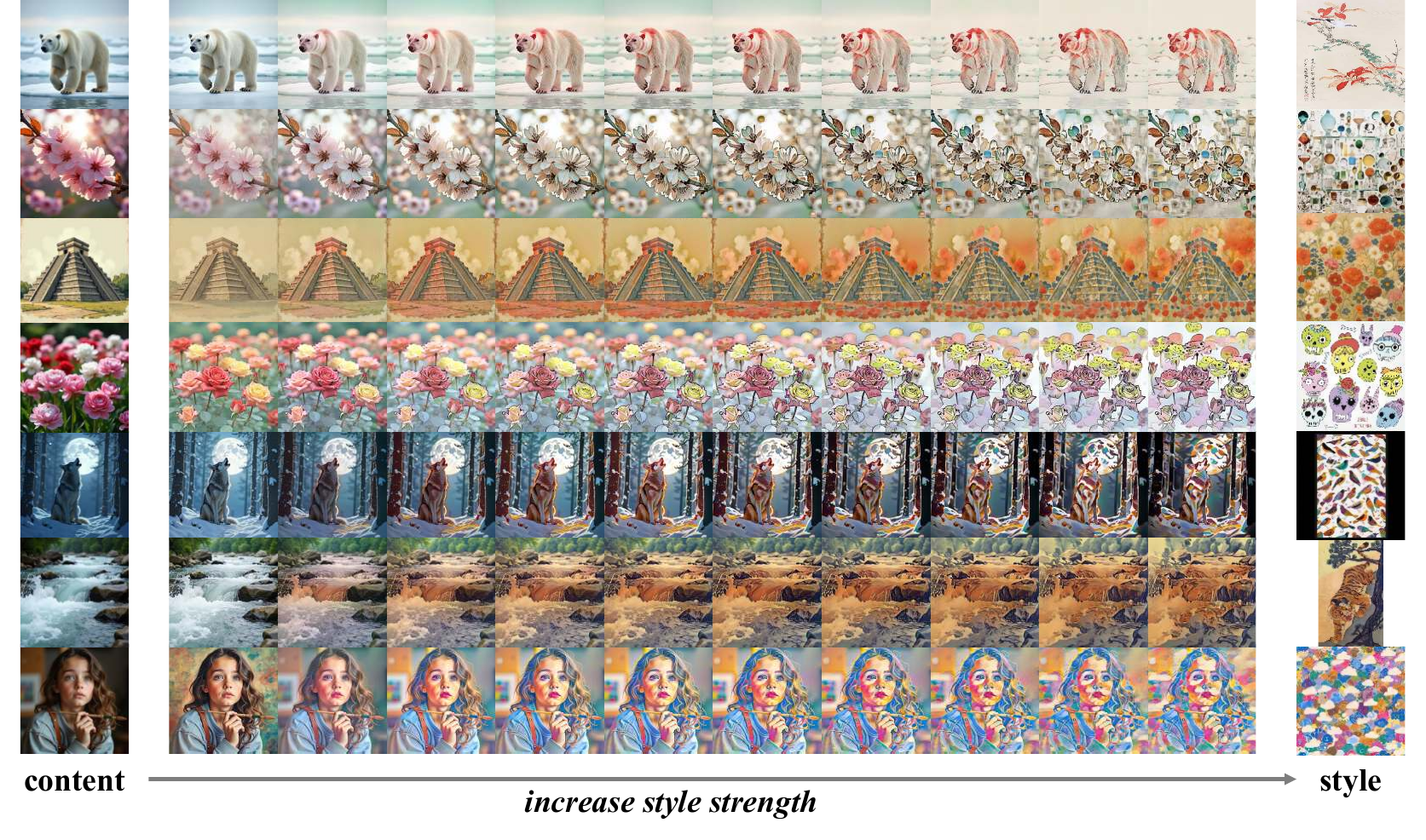}
    \caption{Continuous stylization with smooth transitions. The proposed method transfers styles with explicit stylization control strengths, achieving controllable stylization while preserving content layouts and local style patterns along smooth transition paths.}
    \label{fig:teaser}
\end{figure}

\begin{figure}[t]
  \centering
    \includegraphics[width=\linewidth]{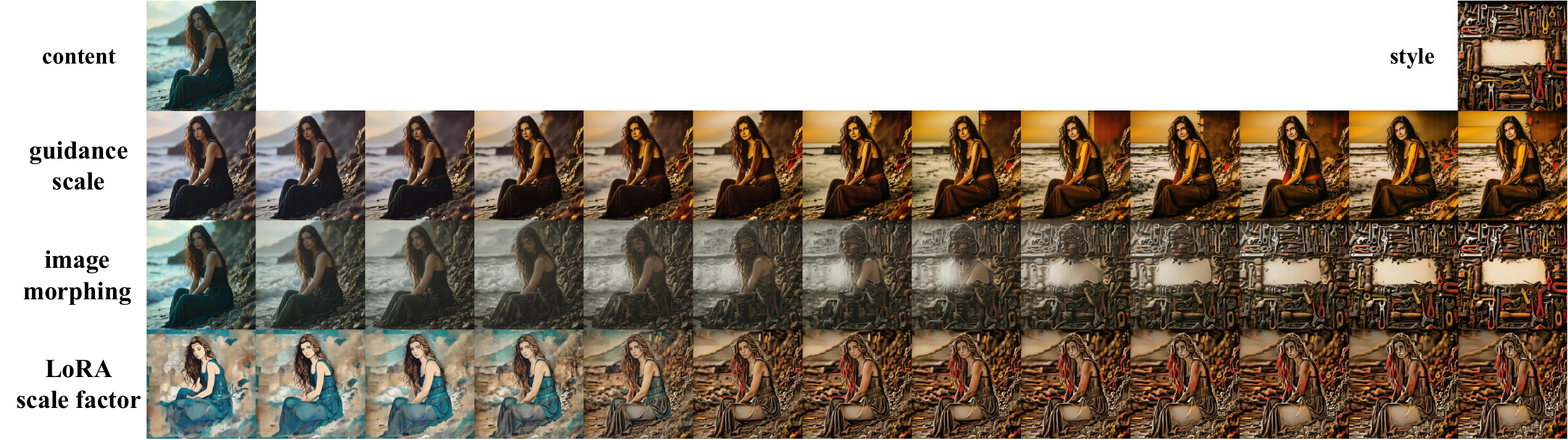}
    \caption{The limitations of current models for continuous stylization with smooth transition control. \textit{Top}: Style transfer models~\cite{stylessp} with guidance scales neglect style patterns and change the content. \textit{Middle}: Morphing-based methods provide a smooth transition but fail to transfer styles to content images. \textit{Bottom}: Scaling a tuned LoRA by adjusting $\frac{\alpha}{r}$ fails to maintain content consistency in the early stages.}
    \label{fig:problems}
\end{figure}

However, most existing methods are constrained to mapping inputs to a fixed stylization strength, which often leads to a reduction in distributional diversity and a failure to capture subtle stylistic patterns. While adjusting guidance scales~\cite{cfg} can partially mitigate these issues, this implicit tuning mechanism typically yields suboptimal results and fails to produce smooth stylization transitions.
Although recent advances in LoRA merging and adapter scaling~\cite{zhong2024multi, chen2025consislora, shenaj2025lora, liu2025unziplora} offer some degree of control, simply manipulating the normalization factors of stylization adapters is inherently unreliable for generating high-fidelity, continuous transition curves.
Alternatively, image morphing techniques~\cite{freemorph, diffmorpher} attempt to create transition paths between images. However, these methods often transform content instances from one to another, significantly compromising stylization performance due to their strong constraints on smoothness.
These persistent challenges are visualized in Figure~\ref{fig:problems}.

Based on current research, a robust methodology for achieving controllable stylization while balancing content preservation and style transfer remains elusive. The limitations of current approaches can be summarized into three core challenges:
(1) \textbf{Content-Style Dilemma}. Existing editing models often struggle to achieve strong stylization effects without compromising the semantic structure of the content images. This manifests as either under-stylized results that lack style patterns or over-stylized outputs where the content becomes unrecognizable.
(2) \textbf{Unreliable Style Control}. Current methods typically map inputs to a fixed, discrete stylization strength. While implicit control methods (\emph{e.g.}, CFG, LoRA) can be applied, these mechanisms are often unreliable and fail to provide predictable and high-fidelity control over the strength of style transitions.
(3) \textbf{Non-Smooth Transition Path}. Although interpolation or morphing techniques can generate intermediate images, they are generally optimized for instance-to-instance blending rather than style-strength-aware transitions. Consequently, they fail to produce smooth stylization transition curves.

To address these challenges, we propose a two-stage training strategy with a strength-aware spline interpolation framework, effectively bridging the gap between local editing and controllable continuous stylization.
In the first stage, the model is tuned to achieve high-fidelity endpoint stylization, focusing on reconstructing content semantics while simultaneously capturing local stylistic patterns and global color palettes. Based on these priors, the second stage introduces a set of anchor projectors that map stylization strengths into a low-rank parameter space. Finally, by formulating a strength-aware spline, we extend the capability of these projectors to enable an explicit and robust control mechanism. In summary, the contributions of this work are three-fold:
\begin{enumerate}
  \item \textbf{Stylization Improvements}. We design a specialized tuning pipeline that empowers image editing models to extract subtle stylistic patterns while maintaining content semantic integrity.
  \item \textbf{Controllable Strengths}. We introduce low-rank anchor projectors that inject stylization strength directly into the latent parameter space, replacing unreliable implicit guidance with an explicit, controllable mechanism.
  \item \textbf{Smooth Trajectories}. We achieve continuous and smooth stylization trajectories by incorporating a strength-aware spline formulation.
\end{enumerate}
Extensive experiments demonstrate that our method exhibits superior stylization performance and robust strength control, consistently preserving both global content structures and fine-grained stylistic details across the entire transition.


\section{Related Work}
\label{sec:related_works}



\textbf{Style Transfer}. With the development of diffusion and flow matching models, style transfer has achieved substantial progress. StyleSSP~\cite{stylessp} improves content preservation through frequency manipulation and employs inversion-based guidance to mitigate content leakage from style references. InstantStyle~\cite{instantstyle,instantstyle_plus} injects style features via attention modulation while preserving content semantics using adapters. Instead of explicitly balancing content and style, StyleDiffusion~\cite{stylediffusion} disentangles style and content representations through a style removal module and learns stylization implicitly. As a representative tuning-based approach, OmniStyle~\cite{omnistyle} constructs a large-scale content--style--stylized dataset and fine-tunes a flow matching model~\cite{flux} to generate aesthetically stylized images.
Despite their effectiveness, these methods provide limited control over stylization strength and often rely on indirect mechanisms such as guidance scales. Consequently, stylization control remains underexplored, potentially leading to unpredictable transfer results in practical applications. In contrast, our method enables continuous and controllable stylization while preserving both content structure and style characteristics.

\textbf{Style Morphing}. Smooth image morphing enables gradual transitions of visual structures and has been widely used in animation and keyframe generation. Although diffusion models can produce high-fidelity images, their unstructured latent spaces make smooth interpolation challenging~\cite{diffmorpher}. \citet{wang2023interpolating} employ textual inversion and perform interpolation in the noisy latent space, followed by denoising to obtain target images. DiffMorpher~\cite{diffmorpher} further introduces linear interpolation over attention and LoRA~\cite{lora} parameters and adopts spherical linear interpolation~\cite{slerp} in the high-dimensional latent space, demonstrating smooth morphing trajectories between images. Building on this line of work, FreeMorph~\cite{freemorph} proposes a training-free framework that incorporates guidance-aware spherical interpolation and customized attention processors.
While these approaches successfully generate transitions between source and target images, they are not designed to explicitly control stylization variations when applied to style transfer. In contrast, our method performs interpolation in a low-rank parameter space, enabling controllable stylization transitions while preserving both content and style characteristics.


\section{Methodology}
\label{sec:methodology}

To improve style transfer and enable controllable stylization transitions, we propose a two-stage, content--style-conditioned fine-tuning method followed by stylization-strength-aware interpolation. As summarized in Figure~\ref{fig:method}, panels (b) and (c) show the two training stages, whereas panel (a) shows inference with an interpolated runtime projector.

\subsection{Preliminaries}
\label{sec:preliminaries}

\textbf{Flow-Matching}. Given the Gaussian distribution $P_{init}=\mathcal{N}(0, I)$ and data distribution $P_{data}$, flow matching aims to construct a marginal vector field $v_t(x)$ ($0 \leq t \leq 1$) mapping a data point $x_0$ drawn from $P_{init}$ to a point $x_1 \sim P_{data}$~\cite{flowmatching} by solving the ODE $\frac{\mathrm{d}}{\mathrm{d}t} X_t = v_t(X_t)$ with initial condition $X_0 = x_0$,
where $x_0 \sim P_{init}$ and $x_t$ follows the marginal probability path $P_t(x_t)$.

Given an arbitrary data point $z \sim P_{data}$ and a Gaussian conditional probability path $P_t(x_t | z)$, we have $P_t(x_t) = \int P_t(x_t|z)P_{data}(z) dz$. With the help of the continuity equation~\cite{flowmatching}, the marginal vector field can be calculated from the conditional field as $v_t(X_t) = \int v_t(X_t|z) \frac{P_t(X_t|z)P_{data}(z)}{P_t(X_t)} dz$.
Then, the tractable conditional flow-matching training objective can be written as $\mathcal{L}_{CFM}(\theta)=\mathbb{E}[\| u^{\theta}_t(X_t) - v_t(X_t|z) \|^2]$. In the Gaussian case, this loss function can be expressed as
\begin{equation}
  \mathcal{L}_{CFM}(\theta)=
  \mathbb{E}\left[\| u^{\theta}_t(\alpha_t \epsilon + \beta_t z) - (\dot{\alpha_t} \epsilon + \dot{\beta_t} z) \|^2\right],
  \label{eq:cfm_loss}
\end{equation}
where $\alpha_t, \beta_t$ are the noise schedulers, and $t\sim \mathcal{U}(0,1), \epsilon\sim \mathcal{N}(0, I), z\sim P_{data}$ are sampled during training.

\subsection{Anchor Style Transfer Dataset}
\label{sec:dataset}

Our method is motivated by the goal of improving stylization performance for image editing models~\cite{qwenimage,seedream,flux} and extending this capability to continuous stylization transitions.
To this end, we construct a new style transfer dataset with stylization anchors~\cite{strotss}.
Specifically, we first generate a collection of style reference images~\cite{flux} with dense texture patterns, which provide priors for learning fundamental style characteristics.
To further enhance style diversity and align the styles with commonly used patterns, we additionally collect a subset of styles from Style30k~\cite{style30k}.
For content images, we randomly sample a subset of images from OmniStyle~\cite{omnistyle}; these images are generated and filtered by expert models~\cite{flux,gpt4,clip,internvl,dinov2}.
Finally, for each content--style pair, we automatically generate five stylization anchors at $s\in\{0.2,0.4,0.6,0.8,1.0\}$ by varying only the content--style trade-off weight, without manual intensity annotation or selection.
For evaluation, we follow the paired evaluation protocol of OmniStyle-150k, separately from the synthetic anchor generation used for training.

\begin{figure*}[t]
  \centering
    \includegraphics[width=\textwidth]{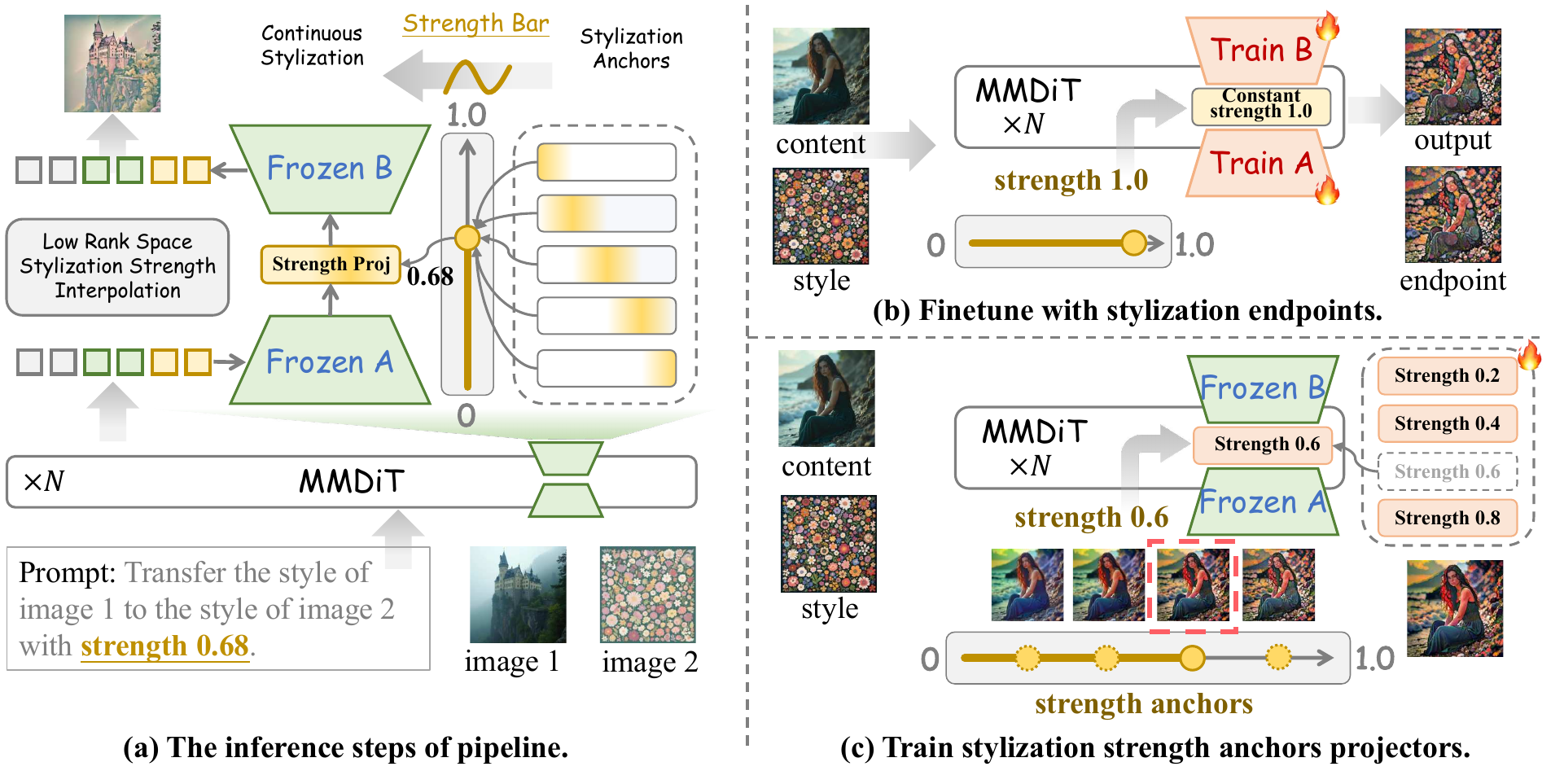}
    \caption{Overview of the proposed two-stage training and inference pipeline. \textit{Left (a)}: during inference, the learned anchor projectors are combined through strength-aware interpolation to construct a runtime projector for a queried strength $s$, which modulates the frozen adapter matrices $A$ and $B$. \textit{Right (b)}: in the first training stage, $A$ and $B$ are optimized using the endpoint at $s=1.0$. \textit{Right (c)}: in the second training stage, $A$ and $B$ are frozen while the strength projectors are trained at discrete stylization anchors.}
    \label{fig:method}
\end{figure*}

\subsection{Stylization Strength Affine Module}

The proposed method aims to improve both basic stylization performance and the ability to perform continuous stylization transitions.
To address the limitations of current image editing models~\cite{qwenimage,seedream,flux}, we adopt a simple yet effective fine-tuning strategy by introducing LoRA~\cite{lora} modules into the attention projection matrices (Q, K, and V)~\cite{transformer} as well as the MLP layers.

Specifically, for a latent representation $x \in \mathbb{R}^d$, LoRA introduces trainable low-rank matrices $A \in \mathbb{R}^{r \times d}$ and $B \in \mathbb{R}^{d \times r}$, where $r \ll d$.
These matrices parameterize a low-rank update to the original weight matrix.
The transformed latent representation can therefore be written as
\begin{equation}
  \tilde{x} = Wx + BAx,
\end{equation}
where $W \in \mathbb{R}^{d \times d}$ denotes the original frozen weight matrix.
Consistent with prior studies and common practices in stylization research~\cite{omnistyle,omnistyle2,omniconsistency,ominicontrol}, incorporating LoRA adapters into a foundation image editing model can significantly improve style transfer performance while keeping the base model parameters frozen.
However, despite these improvements, such a framework still essentially performs a one-to-one mapping between the input content and the target style, which limits its ability to provide fine-grained control over the stylization process.
To enable controllable stylization transitions with a limited number of trainable parameters, training a separate adapter for each target control strength is inefficient.
We observe that these adapters essentially perform the same style transfer task and differ mainly in the stylization strength, while the remaining inputs and outputs remain largely consistent.

Based on this observation, we introduce a low-rank affine transformation that projects latents into different stylization-strength spaces.
Specifically, given a pre-trained style transfer adapter consisting of matrices $A \in \mathbb{R}^{r \times d}$ and $B \in \mathbb{R}^{d \times r}$, we design a lightweight linear module $L \in \mathbb{R}^{r \times r}$ with bias $b \in \mathbb{R}^{r}$ operating in the low-rank space.
The transformed latent representation is defined as
\begin{equation}
  \tilde{x}(s) = Wx + B\left(L\left(\hat{s} \odot Ax\right) + b\right),
  \label{eq:low_rank_linear_lora}
\end{equation}
Here, $s \in [0,1]$ denotes the stylization control strength, and its $r$-dimensional extension is $\hat{s}=s\mathbf{1}_r$, where $\mathbf{1}_r\in\mathbb{R}^r$ is an all-ones vector.
The symbol $\odot$ represents element-wise multiplication.
The stylization strength first scales the low-rank features, after which the full affine projector mixes information across the rank dimensions.

With this design, the trainable parameters consist of a single adapter responsible for the basic style transfer capability, together with a set of lightweight projectors that control the stylization strength.
Specifically, the endpoint adapter is parameterized by $\Delta\theta_0=(A,B)$, while the projector at the $i$-th stylization control level is parameterized by $\Delta\theta_i=(L_i,b_i)$ for $1\le i\le N$.
The model is optimized using a two-stage training strategy with the same linear noise schedule $\alpha_t=t$ and $\beta_t=1-t$ as used in~\cite{rectifiedflow} and described in the preliminaries.
The training objective is defined as
\begin{equation}
  \mathcal{L}(\Delta\theta_i)=
  \mathbb{E}\left[
  \left\|
  u^{\theta+\Delta\theta_i}_t((1-t) z + t \epsilon)
  - (\epsilon - z)
  \right\|^2
  \right],
  \label{eq:target_loss}
\end{equation}
where $t$ denotes the flow-matching time step.

As shown in Figure~\ref{fig:method}(b), the first stage trains the endpoint adapter $\Delta\theta_0=(A,B)$ at $s=1.0$ to establish image-conditioned style transfer capability. In the second stage shown in Figure~\ref{fig:method}(c), $A$ and $B$ are frozen while the strength projectors $\Delta\theta_i$ are trained at the discrete stylization anchors.
Experiments and ablation studies demonstrate that the proposed module is effective for controllable stylization.

\subsection{Interpolation in Low-Rank Space}

We seek continuous stylization control from projectors learned only at the $N$ discrete anchor strengths. We treat the learned projectors $\{\Delta\theta_i=(L_i,b_i)\}_{i=1}^{N}$ as B-spline control points and interpolate them directly in the low-rank parameter space.

For a B-spline of degree $k$, let $\mathcal{U}=\{u_j\}_{j=1}^{N+k+1}$ be a non-decreasing knot vector. The basis functions are defined by the Cox--de Boor recursion
\begin{align}
  B_{i,0}(s) &=
  \begin{cases}
    1, & u_i \le s < u_{i+1}, \\
    0, & \text{otherwise},
  \end{cases} \notag \\
  B_{i,k}(s) &=
  \frac{s-u_i}{u_{i+k}-u_i} B_{i,k-1}(s) \notag \\
  &\quad + \frac{u_{i+k+1}-s}{u_{i+k+1}-u_{i+1}} B_{i+1,k-1}(s),
  \quad k \ge 1,
  \label{eq:bspline2}
\end{align}
where $1\le i\le N$ and a term with a zero denominator is defined as zero. Thus, the $N$ projector control points require $N+k+1$ knots. At a queried strength $s\in[0,1]$, the runtime projector is
\begin{equation}
  \boldsymbol{\Theta}(s)=(L(s),b(s))
  =\sum_{i=1}^{N} B_{i,k}(s)\,\Delta\theta_i.
  \label{eq:low_rank_linear}
\end{equation}
Because the basis weights vary continuously with $s$, $\boldsymbol{\Theta}(s)$ provides smooth strength control while remaining in the low-rank parameter space. Only the projectors $\Delta\theta_i$ are interpolated; the endpoint adapter $\Delta\theta_0=(A,B)$ remains frozen.

At inference, as shown in Figure~\ref{fig:method}(a), we evaluate the basis functions at the queried $s$, construct $\boldsymbol{\Theta}(s)$ using Eq.~\eqref{eq:low_rank_linear}, and combine it with the frozen endpoint adapter to generate the stylized image without further optimization.
Both quantitative and qualitative experimental results demonstrate that the proposed method maintains high stylization fidelity along the entire transition path.


\begin{figure}[t]
  \centering
    \includegraphics[width=\linewidth]{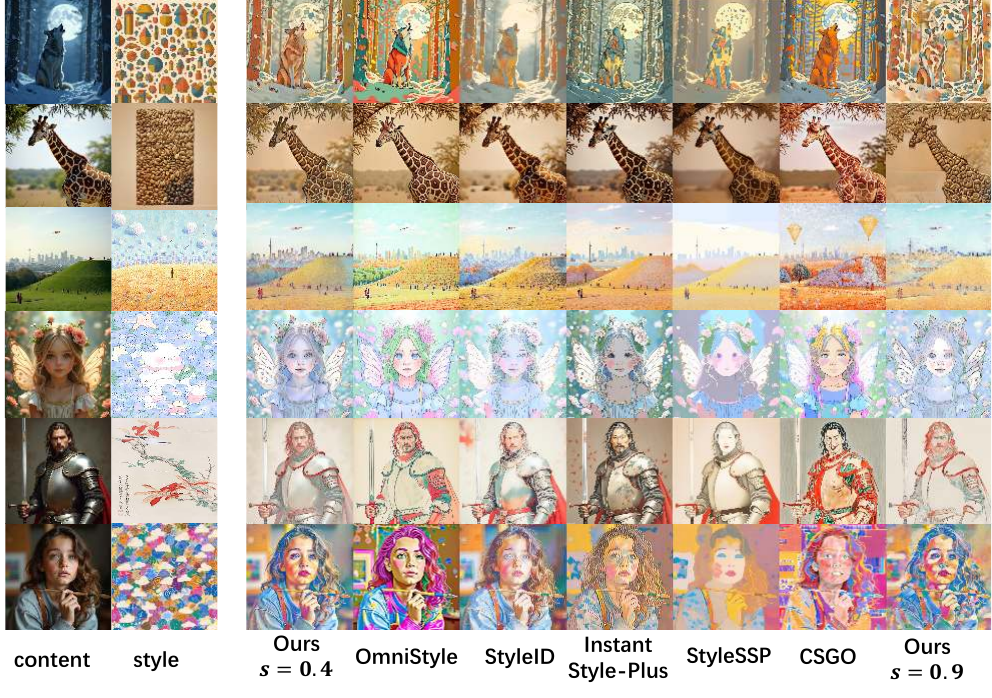}
    \caption{Stylization comparisons between our method and the baselines. We show the default strength $s=0.4$ and \textit{additionally} report $s=0.9$ to evaluate stronger stylization. Zoom in for better visualization.}
    \label{fig:stylization_baseline}
\end{figure}

\section{Experiments}
\label{sec:experiments}

Experiments are conducted using the image editing model QwenImage~\cite{qwenimage} unless otherwise specified.
The parameter-free optimizer Prodigy~\cite{prodigy} is applied with a learning rate of $1.0$.
All experiments are conducted on a single NVIDIA A100 GPU with 80 GB of memory.

\subsection{Metrics}


\textbf{Stylization}. We mainly use FID~\cite{fid}, LPIPS~\cite{lpips}, and ArtFID~\cite{artfid}. FID is used for both style learning and content preservation, denoted as \text{s-FID} and \text{c-FID}, respectively. LPIPS measures the degree of content preservation between stylized images and content images. $\text{ArtFID}=(1+\text{s-FID}) \cdot (1+\text{LPIPS})$ is a comprehensive metric for both content and style transfer capabilities. Meanwhile, we introduce the CLIP image score~\cite{clipscore} for content semantic alignment and style loss~\cite{vgg} ($10^{-4}$) for style alignment, providing a comprehensive view of content and style quality.

\begin{table*}[t]
  \centering
  \begin{tabular*}{\textwidth}{@{\extracolsep{\fill}} l|cccccc @{}}
    \toprule
    Methods & c-FID $\downarrow$ & s-FID $\downarrow$ & LPIPS $\downarrow$ & ArtFID $\downarrow$ & CLIP-I $\uparrow$ & SL $\downarrow$ \\
    \midrule
    StyleID~\cite{styleid} & $118.91$ & $188.48$ & $0.4517$ & $275.06$ & $0.8671$ & $6.0182$ \\
    StyleSSP~\cite{stylessp} & $109.10$ & $186.64$ & $0.4787$ & $277.47$ & $0.8553$ & $6.0903$ \\
    CSGO~\cite{csgo} & $140.77$ & $174.50$ & $0.6008$ & $280.94$ & $0.7954$ & $9.0661$ \\
    InstantStyle-Plus~\cite{instantstyle_plus} & $86.43$ & $202.21$ & $0.3486$ & $274.05$ & $0.9115$ & $7.4530$ \\
    \midrule
    FLUX+OmniStyle~\cite{omnistyle} & $123.35$ & $181.90$ & $0.5404$ & $281.74$ & $0.8546$ & $ 6.9905$ \\
    FLUX+Ours & $91.32$ & $188.09$ & $0.3448$ & $254.29$ & $0.9030$ & $ \textbf{4.6925}$ \\
    \midrule
    QwenImage~\cite{qwenimage} & $125.91$ & $165.74$ & $0.5881$ & $264.81$ & $0.7883$ & $7.4404$ \\
    QwenImage+Ours & $\mathbf{65.95}$ & $\mathbf{165.67}$ & $\mathbf{0.3246}$ & $\mathbf{220.79}$ & $\mathbf{0.9241}$ & $\underline{4.9693}$ \\
  \bottomrule
  \end{tabular*}
  \caption{Quantitative comparisons of stylization performance.}
  \label{tab:baseline_style}
\end{table*}

\begin{table}[t]
  \centering
  \small
  \setlength{\tabcolsep}{3pt}
  \begin{tabular}{@{} l|ccccc @{}}
    \toprule
    & PPL $\downarrow$ & SPL $\downarrow$ & LPIPS $\downarrow$ & SL $\downarrow$ & ArtFID $\downarrow$ \\
    \midrule
    FreeMorph & $2.7632$ & $20.5159$ & $0.4946$ & $6.5588$ & $446.06$ \\
    DiffMorpher & $0.7984$ & $5.0645$ & $0.4091$ & $3.4182$ & $412.57$ \\
    Ours & $\mathbf{0.4959}$ & $\mathbf{4.7162}$ & $\mathbf{0.3530}$ & $\mathbf{3.1354}$ & $\mathbf{408.37}$ \\
  \bottomrule
  \end{tabular}
  \caption{Quantitative comparisons for style morphing.}
  \label{tab:baseline_morph}
\end{table}

\textbf{Smoothness}. We use perceptual path length (PPL)~\cite{ppl} to evaluate the smoothness of the overall generated transition paths. PPL is the sum of perceptual losses~\cite{lpips} between adjacent images along the stylization paths. We then introduce a new metric to better reflect stylization transition smoothness, which we call stylization path length (SPL). Analogous to PPL, SPL focuses on variations in style along the transition path. Specifically, given a stylization transition path $\mathcal{P}=\{I^{cs}_i\}_{i=1}^n$, SPL is calculated using Eq.~\eqref{eq:spl}
\begin{equation}
  \text{SPL}(\mathcal{P}, I_S) = \sum_{i=2}^{n} | \text{SL}(I^{cs}_i, I_S) - \text{SL}(I^{cs}_{i-1}, I_S) |
  \label{eq:spl}
\end{equation}
where $I_S$ is the style reference and $\text{SL}(\cdot)$ is the style loss~\cite{vgg}($10^{-4}$).

\subsection{Qualitative Comparisons}
\label{sec:qualitative}

\textbf{Stylization}. Figure~\ref{fig:stylization_baseline} presents qualitative comparisons between our method and existing style transfer approaches~\cite{csgo,stylessp,styleid,omnistyle,instantstyle_plus}.
We report results at stylization strengths $s=0.4$ and $s=0.9$ to evaluate both moderate and strong stylization.
Compared with the baselines, our method preserves image content more effectively even under strong stylization.
Meanwhile, the generated images maintain global color palettes consistent with the reference styles and progressively reveal richer local style patterns as the stylization strength increases.
For instance, in the last row, the reference style image contains butterfly motifs as its dominant pattern.
While most baselines only transfer coarse attributes such as color and brush strokes, our method successfully incorporates the butterfly pattern into the generated image.
These results demonstrate improved stylization fidelity while maintaining structural consistency with the content image.

\begin{figure}[t]
  \centering
    \includegraphics[width=\linewidth]{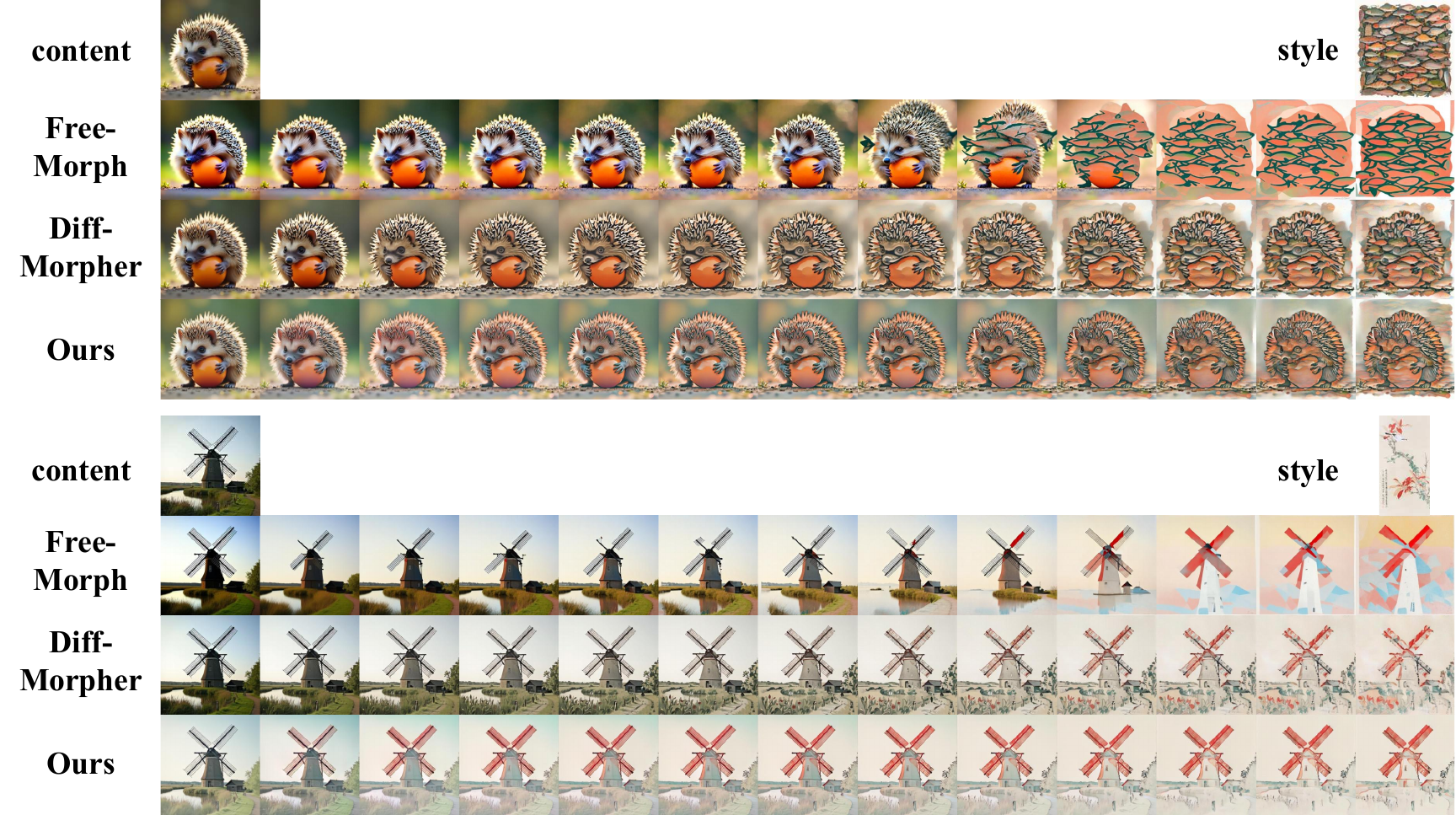}
    \caption{The smoothness of continuous stylization transitions for our method and the baselines. Zoom in for better visualization. Though some morphing methods can transfer the reference styles when \textit{conditioned on ground-truth endpoints}, they still fail to preserve content along the transition path.}
    \label{fig:continuous_baseline}
\end{figure}

\textbf{Smoothness}. Morphing models~\cite{diffmorpher,freemorph} require both start and end images as inputs; however, in stylization scenarios, the end image is not naturally available.
To enable comparison, we use the ground-truth stylized images from our dataset as endpoint images for the morphing models and evaluate the generated intermediate frames without requiring ground-truth intermediate frames.
As shown in Figure~\ref{fig:continuous_baseline}, our method preserves the content structure along the transition path while progressively injecting global style colors and local patterns.
Although both our method and DiffMorpher produce visually smooth transitions, our approach transfers stylistic patterns more effectively even at low stylization strengths.
Overall, the proposed method achieves both strong style transfer capability and smooth stylization transitions.

\subsection{Stylization Quantitative Comparisons}

To further evaluate the stylization performance of the proposed method, Table~\ref{tab:baseline_style} reports quantitative comparisons with style transfer models.
For content preservation, our method achieves the lowest \text{LPIPS} ($0.3246$) and content \text{FID} ($65.95$) among all compared methods.
Meanwhile, the model also preserves semantic consistency with the content images, as reflected by the highest \text{CLIP} image score ($0.9241$).
Regarding style learning capability, both the style \text{FID} and style loss remain at relatively low levels, indicating strong stylistic similarity between the generated images and the reference styles.
For overall stylization quality, benefiting from both strong content fidelity and effective style modeling, our method achieves the lowest \text{ArtFID} ($220.79$), demonstrating competitive performance.

These quantitative results further support the qualitative observations in Figure~\ref{fig:stylization_baseline} and are consistent with the qualitative analysis above.
Overall, both qualitative and quantitative evaluations indicate that the proposed method achieves a favorable balance between accurate content preservation and effective style pattern transfer.

\begin{figure}[t]
  \centering
  \begin{subfigure}{\linewidth}
    \includegraphics[width=\linewidth]{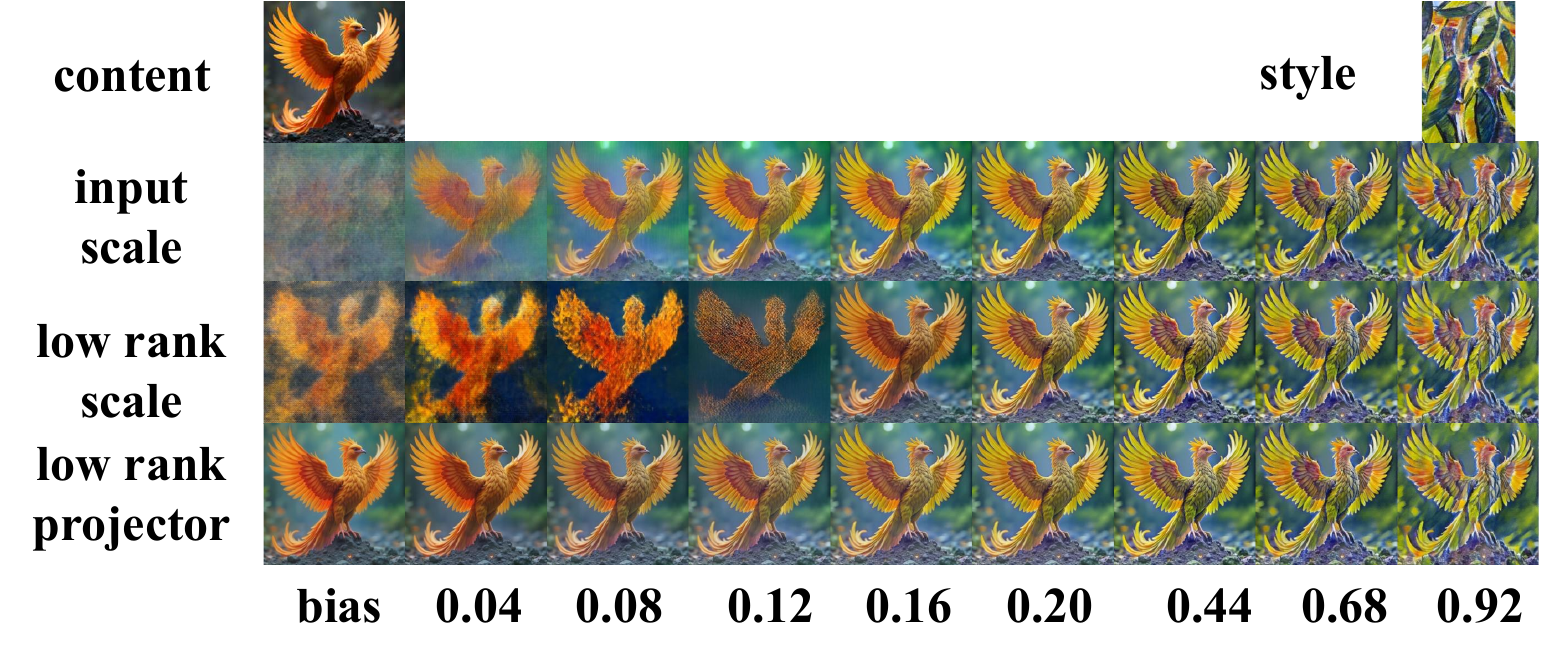}
    \caption{Ablation for projector types.}
    \label{fig:ablation_interp}
  \end{subfigure}
  \hfill
  \begin{subfigure}{\linewidth}
    \includegraphics[width=\linewidth]{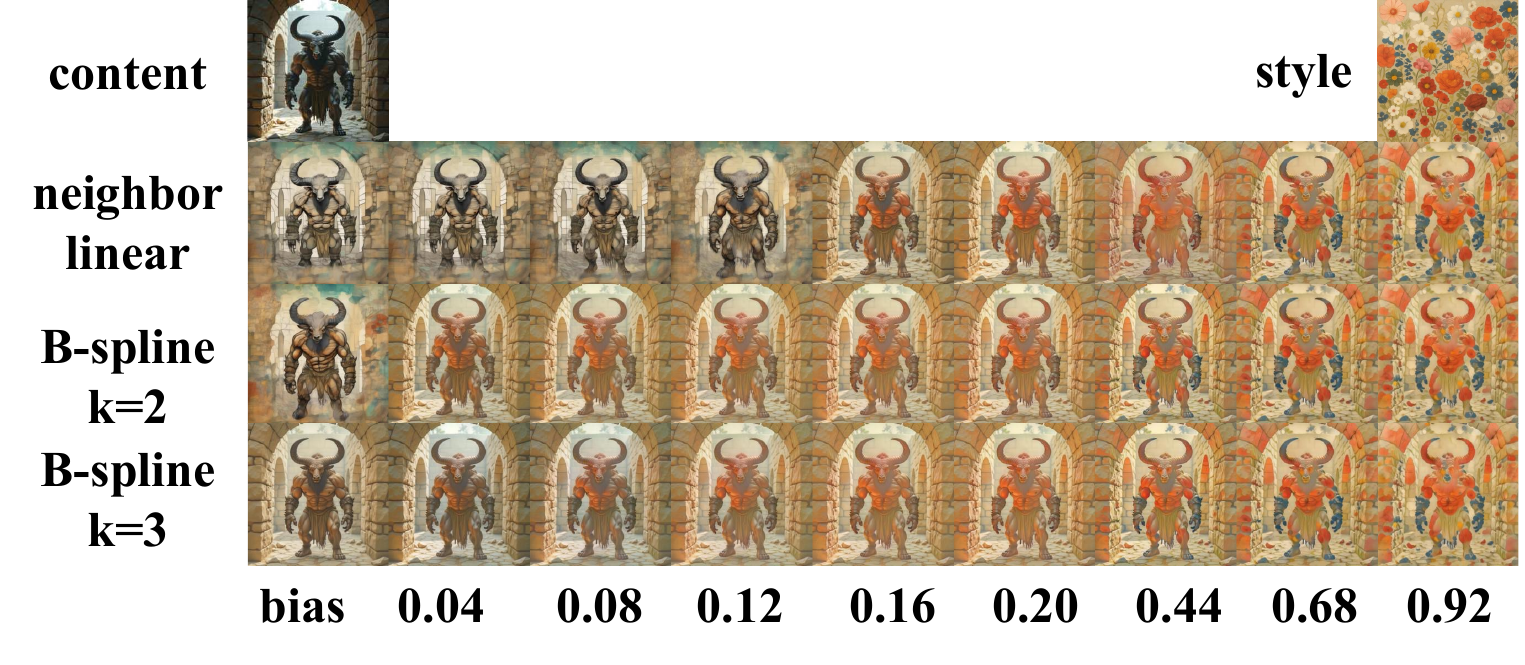}
    \caption{Ablation for interpolation degree.}
    \label{fig:ablation_order}
  \end{subfigure}
  \caption{Ablation studies on projector types (top) and interpolation degree (bottom). Note that the sampled strengths increase nonlinearly.}
  \label{fig:abl-interp-degree}
\end{figure}

\subsection{Style Morphing Quantitative Comparisons}

Our method further produces highly smooth stylization transitions compared with morphing-based models.
Note that morphing methods require both the content image and the \textit{ground-truth} stylized image as inputs.
As shown in Table~\ref{tab:baseline_morph}, FreeMorph~\cite{freemorph} struggles to perform effective style transfer even when conditioned on ground-truth endpoints, resulting in significantly higher metric values.
Compared with DiffMorpher~\cite{diffmorpher}, our method achieves the best \text{PPL} ($0.4959$), indicating stronger content preservation and smoother structural transitions.
Moreover, the relatively low $\text{SPL}=4.7162$ demonstrates smoother style injection throughout the transition process.
These observations are consistent with the qualitative results in Figure~\ref{fig:continuous_baseline}, where our method introduces style patterns at early stages and produces stable transitions from content images to stylized outputs.
Overall, by controlling stylization strength, our approach generates high-fidelity images while maintaining smooth transition trajectories, providing an effective solution for controllable style transfer.

\begin{figure}[t]
  \centering
  \begin{subfigure}{0.49\linewidth}
    \includegraphics[width=\linewidth]{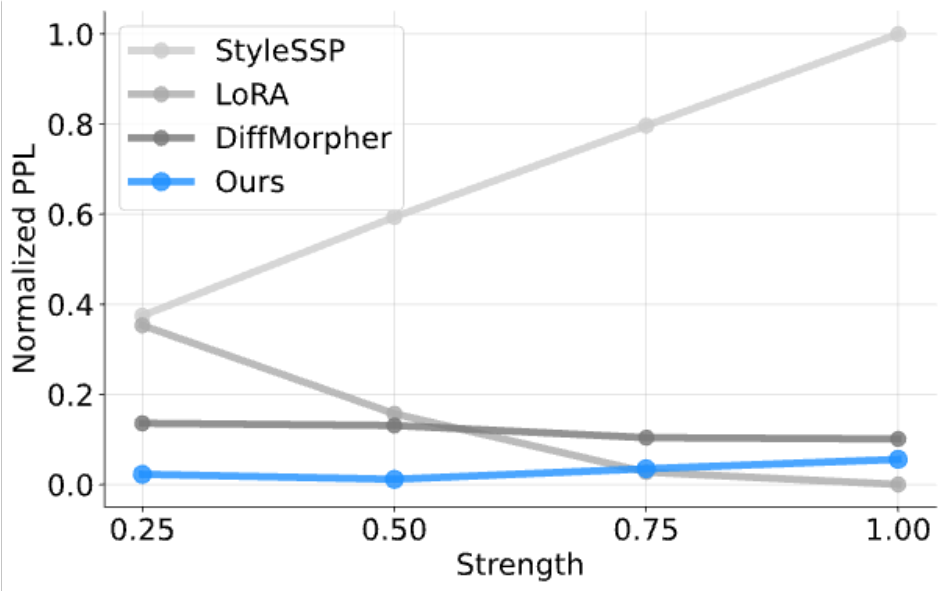}
    \label{fig:ppl_line}
  \end{subfigure}
  \hfill
  \begin{subfigure}{0.49\linewidth}
    \includegraphics[width=\linewidth]{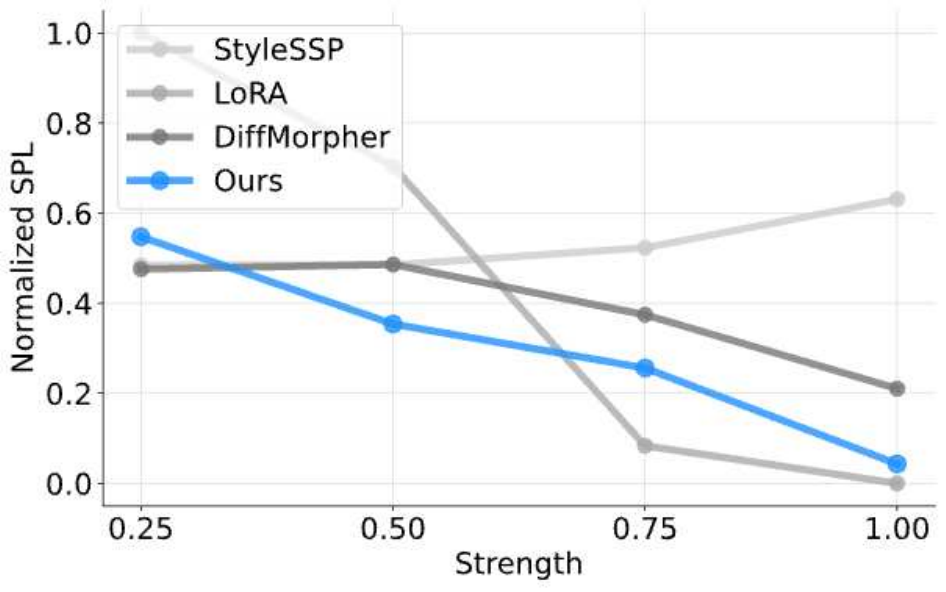}
    \label{fig:spl_line}
  \end{subfigure}
  \caption{PPL (\textit{Left}) and SPL (\textit{Right}) along the stylization transition paths for the guidance-scale, LoRA-scaling, morphing, and proposed methods.}
  \label{fig:ppl_spl_lines}
\end{figure}

We further compare different transition strategies (guidance-scale control, LoRA scaling, morphing, and our method) in Figure~\ref{fig:ppl_spl_lines}.
As illustrated in the left panel of Figure~\ref{fig:ppl_spl_lines}, our method maintains consistently low and stable \text{PPL} values, reflecting strong content preservation, whereas the guidance-scale and LoRA-scaling strategies fail to retain fine structural details.
For style transition smoothness, the right panel of Figure~\ref{fig:ppl_spl_lines} shows that the guidance-scale approach deviates from the reference style, while the LoRA-scaling strategy exhibits abrupt changes in the middle stages, leading to suboptimal stylization control.

\subsection{Ablation Study}

\begin{table}[t]
  \centering
  \setlength{\tabcolsep}{4pt}
  \begin{tabular}{@{}l|ccccc@{}}
    \toprule
    & PPL $\downarrow$ & SPL $\downarrow$ & LPIPS $\downarrow$ & SL $\downarrow$ & ArtFID $\downarrow$ \\
    \midrule
    Eq.~\eqref{eq:ablation_interp1} & $1.2863$ & $7.1156$ & $0.3952$ & $3.2280$ & $439.16$ \\
    Eq.~\eqref{eq:ablation_interp2} & $2.3582$ & $8.4097$ & $0.4400$ & $3.3768$ & $459.16$ \\
    Ours & $\mathbf{0.4959}$ & $\mathbf{4.7162}$ & $\mathbf{0.3530}$ & $\mathbf{3.1354}$ & $\mathbf{408.37}$ \\
  \bottomrule
  \end{tabular}
  \caption{Ablation study for different projectors. Note that our method in Eq.~\eqref{eq:low_rank_linear_lora} is denoted as ``low-rank space linear''.}
  \label{tab:ablation_proj}
\end{table}

\begin{table}[t]
  \centering
  \setlength{\tabcolsep}{5pt}
  \begin{tabular}{@{}l|ccccc@{}}
    \toprule
    & PPL $\downarrow$ & SPL $\downarrow$ & LPIPS $\downarrow$ & SL $\downarrow$ & ArtFID $\downarrow$ \\
    \midrule
    $n=1$ & $1.2217$ & $6.8433$ & $0.4926$ & $4.0041$ & $436.58$ \\
    $n=3$ & $1.3639$ & $7.9803$ & $0.4070$ & $3.5348$ & $421.66$ \\
    $n=5$ & $1.1407$ & $6.6244$ & $0.3922$ & $3.1499$ & $416.97$ \\
    $k=2$ & $\underline{0.5233}$ & $\mathbf{4.6132}$ & $\underline{0.3573}$ & $\mathbf{3.1008}$ & $\underline{409.03}$ \\
    $k=3$ & $\mathbf{0.4959}$ & $\underline{4.7161}$ & $\mathbf{0.3531}$ & $\underline{3.1354}$ & $\mathbf{408.37}$ \\
  \bottomrule
  \end{tabular}
  \caption{Ablation study for strength-aware interpolation. The best and second-best results are shown in bold and underlined, respectively. The symbol $n$ denotes nearest-neighbor interpolation with $n$ neighbors, while $k$ denotes the degree of the B-spline.}
  \label{tab:ablation_interp}
\end{table}

To validate the effectiveness of the proposed components, we conduct ablation experiments to compare different settings.

\begin{figure}[t]
  \centering
  \begin{subfigure}{0.49\linewidth}
    \includegraphics[width=\linewidth]{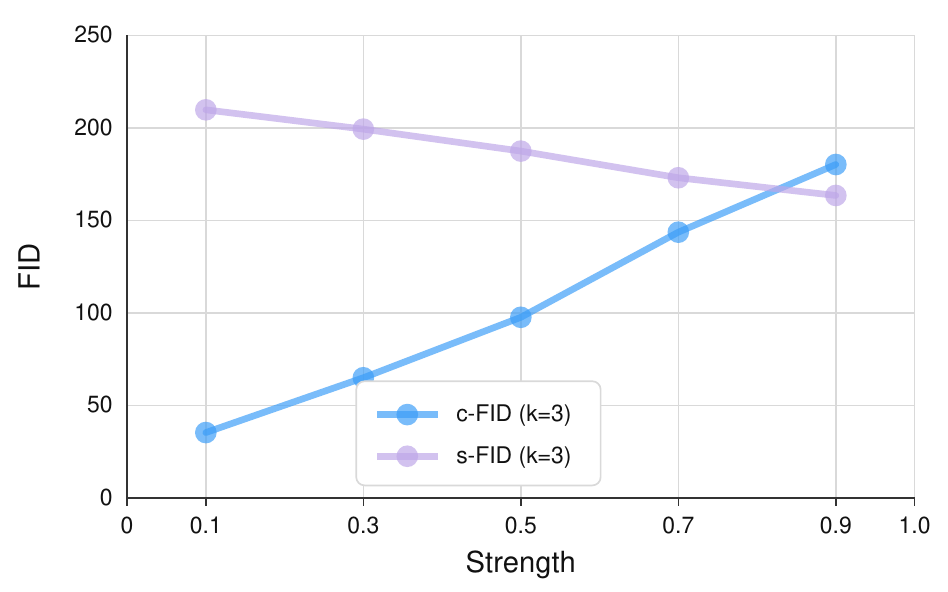}
  \end{subfigure}
  \hfill
  \begin{subfigure}{0.49\linewidth}
    \includegraphics[width=\linewidth]{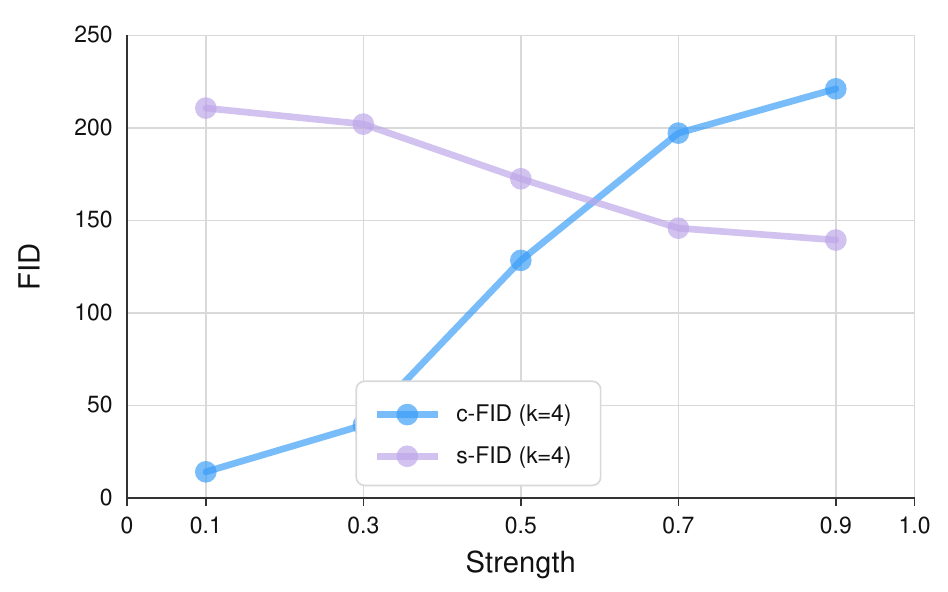}
  \end{subfigure}
  \caption{The c-FID (blue) and s-FID (purple) across sampled stylization strengths for B-spline degrees $k=3$ and $k=4$.}
  \label{fig:abl_k_FID}
\end{figure}

\textbf{Projectors}. Ablations on the projectors aim to illustrate the effectiveness of the low-rank-space linear projector. Inspired by~\cite{lora} and its implementations, we design two other common frameworks in Eqs.~\eqref{eq:ablation_interp1} and~\eqref{eq:ablation_interp2}, which are similar to the proposed module because they all aim to modulate the feature dimension to enable strength-aware capability.
\begin{align}
  \tilde{x}(s) &= Wx + BA (\omega_1 \odot s \odot x + b_1) \label{eq:ablation_interp1} \\
  \tilde{x}(s) &= Wx + B (\omega_2 \odot s \odot Ax + b_2)
  \label{eq:ablation_interp2}
\end{align}
Here, $\omega_1,b_1 \in \mathbb{R}^d$ and $\omega_2,b_{\omega_2},b \in \mathbb{R}^r$ $(d \gg r)$ are learnable vectors, with $b_2:=b_{\omega_2}\odot Ax+b$. Unlike these direct vector-scaling baselines, our formulation applies the full matrix projector $L$ to the strength-scaled low-rank features, enabling interactions across rank dimensions while supporting spline interpolation in the projector parameter space.

As shown in Table~\ref{tab:ablation_proj}, employing learnable scalars in the low-rank space leads to the weakest style transfer performance and transition smoothness. Applying scalars in the input space yields relatively improved results but remains inferior to our linear projector design. The qualitative comparisons in Figure~\ref{fig:abl-interp-degree} further support these observations. Specifically, scalar-based methods in both the high-dimensional input space and low-rank space fail to preserve content structure and layout at early stages where stylization control strengths are small. In contrast, our method reconstructs content and layout while progressively injecting style features, demonstrating greater robustness to fine-grained strength variations, which validates the effectiveness of linear projectors in the low-rank space.

\textbf{Interpolation}. The choice of interpolation strategy significantly affects stylization transition quality. As reported in Table~\ref{tab:ablation_interp}, we compare linear interpolation with spline approaches of different degrees. Compared with the commonly used piecewise linear interpolation between $n$ adjacent parameters~\cite{lora}, spline methods consistently achieve superior performance. As illustrated in the first row of Figure~\ref{fig:abl-interp-degree}, linear interpolation fails to preserve content at early stages near the stylization boundary, whereas spline methods exhibit faster convergence toward stable stylization. Regarding spline degree, the methods show comparable performance overall, with $k=2$ achieving slightly better style transfer metrics. However, this setting demonstrates reduced robustness and occasionally produces images with lower content fidelity, as shown in the second row of Figure~\ref{fig:abl-interp-degree}.

\subsection{Additional Analysis on Strength Control}
The purpose of this analysis is to verify whether the queried strength $s$ provides predictable control over the content--style trade-off \textit{beyond} the discrete anchors used for training and to compare different spline degrees. We therefore evaluate degrees $k=3$ and $k=4$ across a range of stylization strengths.
As shown in Figure~\ref{fig:abl_k_FID}, for both spline degrees, c-FID increases monotonically and s-FID decreases monotonically as $s$ grows, with no reversal between adjacent queries. Thus, increasing $s$ consistently relaxes content preservation while strengthening style matching, rather than producing an irregular response between the learned anchors. However, $k=3$ yields a more gradual trade-off, whereas $k=4$ changes more sharply in the middle range and reaches stronger style matching at high strengths at the cost of larger content deviation.
These results demonstrate that $s$ acts as an interpretable control variable and that the proposed interpolation generalizes discrete anchor projectors to unseen strengths. Compared with $k=4$, $k=3$ avoids the sharp increase in c-FID over the middle and high strength ranges while still reducing s-FID steadily. We therefore select $k=3$ as the default degree because it provides a more gradual and balanced control response across the full stylization path; $k=4$ favors stronger style matching at high strengths but incurs substantially larger content deviation.


\section{Conclusions}
\label{sec:conclusion}

This paper presents a controllable continuous stylization framework for generating smooth style transition paths in image editing models. To overcome the limited stylization control of existing methods, we adopt an endpoint stylization tuning strategy and introduce learnable projectors in low-rank spaces to better capture style strength variations. Furthermore, strength-aware interpolation is performed in the low-rank space to ensure smooth transitions along the stylization trajectory. Extensive experiments and ablation studies demonstrate that the proposed method achieves strong style transfer fidelity, effective style pattern learning, and stable continuous stylization with smooth transition dynamics.

\appendix
\twocolumn[
  \centering
  {\LARGE\bfseries Supplementary Materials\par}
  \vspace{1em}
]
This standalone supplementary material contains the following additional information and experiments:
\begin{itemize}
  \item implementation details for reproducing the method;
  \item user studies of stylization quality and transition smoothness;
  \item algorithms for the two training stages and runtime interpolation;
  \item fixed-content experiments that isolate robustness to the style reference;
  \item additional comparisons with stylization baselines; and
  \item additional continuous-strength results.
\end{itemize}

\section{Implementation Details}

We use QwenImage~\cite{qwenimage} as the default image-editing backbone. All training is conducted on a single NVIDIA A100 GPU with 80~GB of memory at a resolution of $512 \times 512$. The first stage optimizes the endpoint LoRA adapter for $2500$ steps at stylization strength $s=1.0$. The second stage freezes the backbone and endpoint adapter and optimizes the strength projectors for $500$ steps using five anchor strengths, $\{0.2,0.4,0.6,0.8,1.0\}$. Both stages use the Prodigy optimizer~\cite{prodigy} with a learning rate of $1.0$ and zero weight decay.

At inference, our default sampler uses $16$ steps. The NaViT-based visual encoder~\cite{navit} also permits inputs whose resolutions differ from the $512 \times 512$ training resolution. Competing methods use the inference settings recommended by their respective official implementations.

For strength-aware interpolation, we use SciPy's interpolating B-spline routine~\cite{scipy} with the anchor strengths as interpolation nodes, the stacked projector tensors as values, spline degree $k=3$, and \verb|bc_type=None|. The resulting not-a-knot interpolating B-spline passes through every learned anchor projector. Consequently, querying an anchor strength exactly recovers its tuned projector, while querying an intermediate strength constructs a new projector in the same low-rank parameter space.

\section{User Study}

The user study evaluates two perceptual properties that are difficult to capture completely with automated metrics: the overall quality of a stylized output and the perceived smoothness of a stylization trajectory.
For stylization quality, the survey compares our method with OmniStyle~\cite{omnistyle}, StyleID~\cite{styleid}, InstantStyle-Plus~\cite{instantstyle_plus}, StyleSSP~\cite{stylessp}, and CSGO~\cite{csgo} under shared content--style inputs. For transition smoothness, it compares our strength-controlled trajectories with DiffMorpher~\cite{diffmorpher} and FreeMorph~\cite{freemorph}. Because these morphing methods require both endpoints, they are additionally given the target stylized images from the evaluation pairs. Tables~\ref{tab:style-preference} and~\ref{tab:continuous-preference} report normalized shares of the collected preference votes.

\begin{table*}[t]
    \centering
    \small
    \setlength{\tabcolsep}{5pt}
    \begin{tabular}{@{}l| *{6}{c}@{}}
      \toprule
      & OmniStyle & StyleID & InstantStyle-Plus & StyleSSP
      & CSGO & \textbf{Ours}  \\
      \midrule
      Preference (\%)
      & 12.83 & 14.80 & 19.74 & 15.46
      & 11.84 & \textbf{25.33}  \\
      \bottomrule
    \end{tabular}
    \caption{Normalized user-preference shares for overall stylization quality. Higher is better, and all entries sum to $100\%$.}
    \label{tab:style-preference}
\end{table*}

\begin{table}[t]
    \centering
    \small
    \setlength{\tabcolsep}{6pt}
    \begin{tabular}{@{}l| *{3}{c}@{}}
      \toprule
      & DiffMorpher & FreeMorph & \textbf{Ours}  \\
      \midrule
      Preference (\%)
      & 31.58 & 27.63 & \textbf{40.79}  \\
      \bottomrule
    \end{tabular}
    \caption{Normalized user-preference shares for stylization-transition smoothness. DiffMorpher and FreeMorph are supplied with target stylized endpoints. Higher is better, and all entries sum to $100\%$.}
    \label{tab:continuous-preference}
\end{table}

As shown in Table~\ref{tab:style-preference}, our method receives the largest stylization vote share, $25.33\%$. It exceeds the second-ranked InstantStyle-Plus ($19.74\%$) by $5.59$ percentage points. Table~\ref{tab:continuous-preference} shows a larger advantage for continuous transitions: our method receives $40.79\%$ of the votes, compared with $31.58\%$ for DiffMorpher and $27.63\%$ for FreeMorph. The margin over the strongest morphing baseline is $9.21$ percentage points.
The largest vote share in both surveys indicates that the proposed method's gains are perceptually meaningful. In particular, the transition result is notable because the morphing baselines receive an extra target endpoint, whereas our method generates the trajectory directly from a content image, a style reference, and the queried strengths. The stylization result should be interpreted as a plurality rather than a majority: $25.33\%$ is the highest share among six methods but is below $50\%$.

\section{Algorithms}

Algorithms~\ref{alg:train1} and~\ref{alg:train2} specify the two training stages, and Algorithm~\ref{alg:infer} specifies runtime interpolation. Let $z$ denote a target stylized latent, $\epsilon\sim\mathcal{N}(0,I)$ denote Gaussian noise, and $t\sim\mathcal{U}(0,1)$. Both stages minimize the flow-matching objective with noisy input $z_t=(1-t)z+t\epsilon$ and target velocity $\epsilon-z$. The backbone parameters $\theta$ remain frozen throughout training.

\begin{algorithm}[t]
\caption{Stage 1: endpoint adapter training}
\label{alg:train1}
\small
\begin{algorithmic}[1]
\REQUIRE Frozen backbone $\theta$; endpoint set $\mathcal{D}_{1.0}$
\STATE Initialize LoRA matrices $A$ and $B$
\FOR{$m=1,\ldots,2500$}
  \STATE Sample $(I_c,I_s,z)\sim\mathcal{D}_{1.0}$
  \STATE Sample $t\sim\mathcal{U}(0,1)$ and $\epsilon\sim\mathcal{N}(0,I)$
  \STATE $z_t\leftarrow(1-t)z+t\epsilon$; $v^\star\leftarrow\epsilon-z$
  \STATE $\hat v\leftarrow u_{\theta,A,B}(z_t,t\mid I_c,I_s)$
  \STATE Update $A,B$ using $\|\hat v-v^\star\|_2^2$
\ENDFOR
\RETURN Frozen endpoint adapter $(A,B)$
\end{algorithmic}
\end{algorithm}

\begin{algorithm}[t]
\caption{Stage 2: anchor-projector training}
\label{alg:train2}
\small
\begin{algorithmic}[1]
\REQUIRE Frozen $\theta,A,B$; anchor sets $\{\mathcal{D}_{s_i}\}_{i=1}^{N}$
\STATE Initialize $P_i=(L_i,b_i)$ for every anchor $s_i$
\FOR{$m=1,\ldots,500$}
  \STATE Sample anchor $i$ and $(I_c,I_s,z)\sim\mathcal{D}_{s_i}$
  \STATE Sample $t\sim\mathcal{U}(0,1)$ and $\epsilon\sim\mathcal{N}(0,I)$
  \STATE $z_t\leftarrow(1-t)z+t\epsilon$; $v^\star\leftarrow\epsilon-z$
  \STATE $\hat v\leftarrow u_{\theta,A,B,P_i}(z_t,t\mid I_c,I_s,s_i)$
  \STATE Update only $P_i$ using $\|\hat v-v^\star\|_2^2$
\ENDFOR
\RETURN Anchor projectors $\{P_i\}_{i=1}^{N}$
\end{algorithmic}
\end{algorithm}

\begin{algorithm}[t]
\caption{Inference: construct a runtime projector}
\label{alg:infer}
\small
\begin{algorithmic}[1]
\REQUIRE Query $s$; nodes $X=[0.2,0.4,0.6,0.8,1.0]$;
\REQUIRE \hspace{1.3em}anchor projectors $\{P_i\}_{i=1}^{5}$; frozen $\theta,A,B$
\STATE Group corresponding tensors in $\{P_i\}$ by parameter key
\FOR{each parameter key $q$}
  \STATE $Y_q\leftarrow\operatorname{stack}(P_1[q],\ldots,P_5[q])$
  \STATE $\mathcal{S}_q\leftarrow\operatorname{make\_interp\_spline}(X,Y_q,k=3)$
  \STATE $P(s)[q]\leftarrow\mathcal{S}_q(s)$
\ENDFOR
\STATE Load $P(s)$ alongside the frozen endpoint adapter $(A,B)$
\RETURN Generate the output with the standard $16$-step sampler
\end{algorithmic}
\end{algorithm}

\section{Content Robustness}

\begin{figure*}[t]
  \centering
  \includegraphics[width=\linewidth]{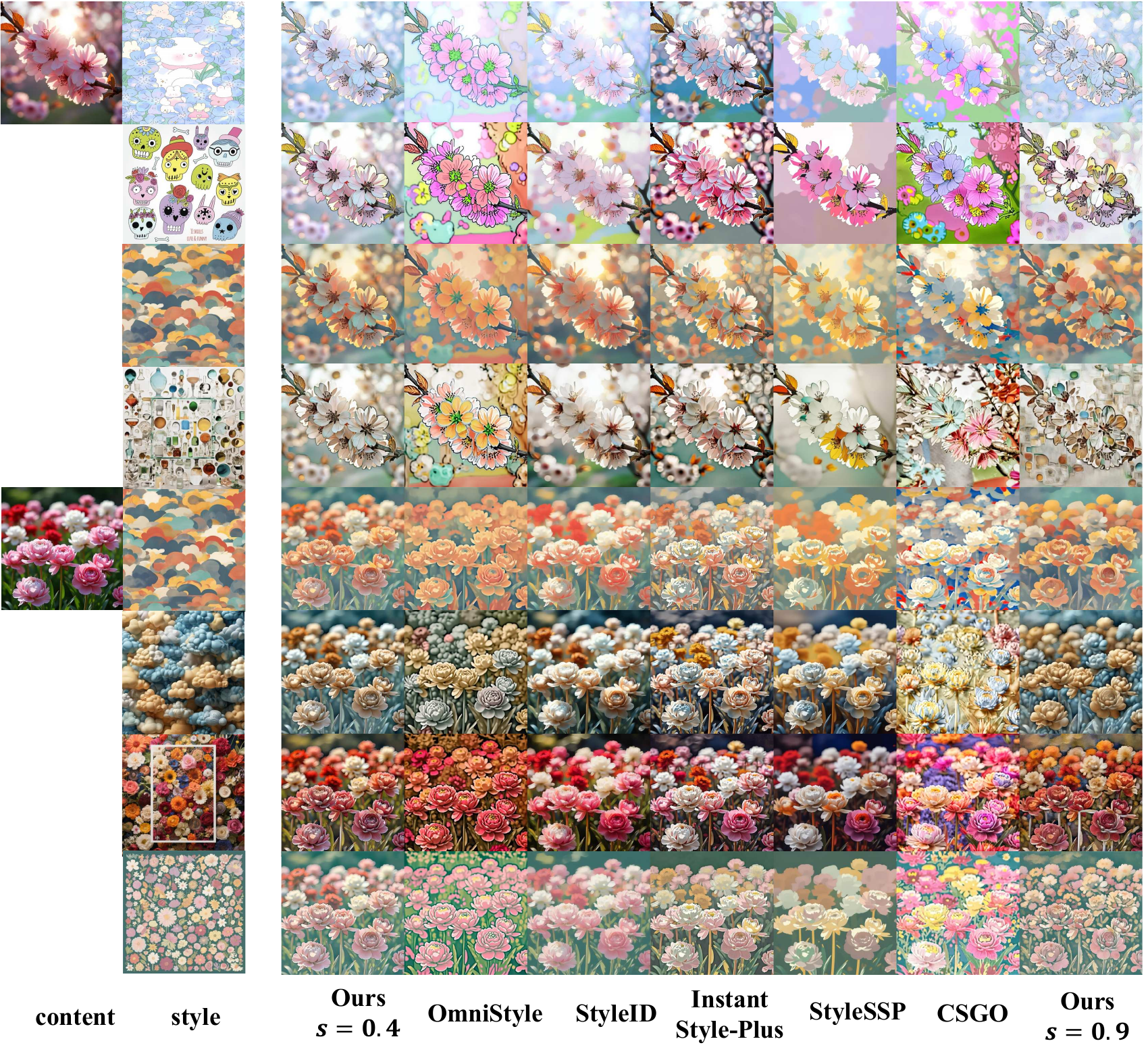}
  \caption{Fixed-content robustness comparisons. Within each of the two content groups, the content image is held fixed while four style references are varied. Columns show the content image, style reference, our method at $s=0.4$, OmniStyle, StyleID, InstantStyle-Plus, StyleSSP, CSGO, and our method at $s=0.9$. Our two settings illustrate the progression from conservative transfer to stronger reference-specific colors and motifs.}
  \label{fig:style_compare2}
\end{figure*}

This experiment isolates sensitivity to the style reference. Holding the content image fixed makes it easier to determine whether a method can alter color, texture, and local motifs without changing the underlying object geometry.
Figure~\ref{fig:style_compare2} contains two fixed-content groups: a close-up of blossoms and a bed of roses. Within each group, the same content is paired with four diverse references, including flat illustrations, dense object collages, and repeated motifs. We compare our outputs at $s=0.4$ and $s=0.9$ with five dedicated stylization methods. Morphing methods are excluded because they require a target stylized endpoint and therefore operate under a different input setting.

At $s=0.4$, our method generally retains the original flower locations, silhouettes, and depth cues while shifting the global palette toward each reference. At $s=0.9$, reference-specific local patterns become more visible: rounded cloud-like regions, outlined cartoon forms, and collage-like object shapes are incorporated without replacing the blossom branches or the spatial extent of the rose bed. Across the baselines, weak transfer often leaves the source photograph nearly unchanged, whereas aggressive transfer can flatten or rearrange the flowers. Our two strength levels expose the intended progression between these extremes.
The consistent behavior across multiple references for each fixed content image indicates that strength control is not tied to a particular content--style pair. The model can increase reference-specific appearance changes while retaining the principal content layout.

\section{More Stylization Comparisons}

\begin{figure*}[t]
  \centering
  \includegraphics[width=\linewidth]{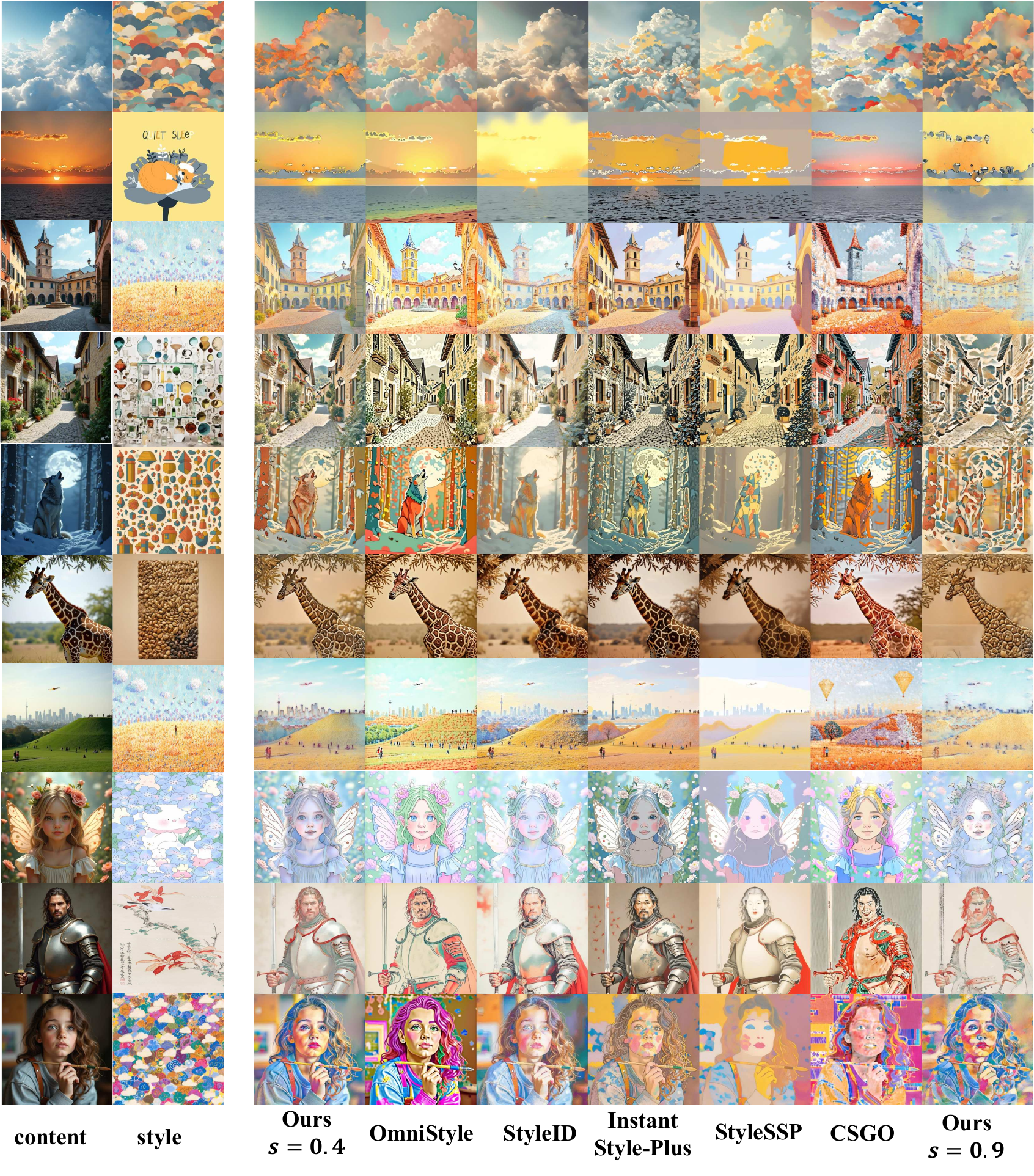}
  \caption{Additional stylization comparisons across diverse content--style pairs. Columns show the content image, style reference, our method at $s=0.4$, OmniStyle, StyleID, InstantStyle-Plus, StyleSSP, CSGO, and our method at $s=0.9$. The moderate setting emphasizes content preservation, whereas the stronger setting introduces more local patterns from the reference.}
  \label{fig:style_compare}
\end{figure*}

This experiment evaluates generalization across varied content categories and style types, and examines whether explicit strength control provides useful outputs at both moderate and strong settings.
Figure~\ref{fig:style_compare} compares the same five stylization baselines with our method at $s=0.4$ and $s=0.9$. The examples cover landscapes, architecture, animals, and portraits, paired with references containing watercolor washes, geometric illustrations, ink drawings, repeated objects, and high-saturation paintings. Every row uses a common content image and style reference for all methods.

The $s=0.4$ outputs favor content preservation: cloud masses, street perspective, animal silhouettes, facial identity, and other scene structures remain recognizable while the reference palette is introduced. Increasing the strength to $s=0.9$ makes local style evidence more explicit, such as coffee-bean-like texture on the giraffe, geometric marks in the forest scene, ink-like contours on the armored figure, and saturated painterly regions on the portrait. Several competing methods either transfer mainly color with limited motif adoption or introduce stronger appearance changes together with noticeable structural drift.
The comparisons show that a single fixed output does not fully characterize stylization quality. The two queried strengths provide distinct and useful operating points, allowing users to trade conservative content preservation for stronger transfer of local style patterns.

\section{More Results}
\begin{figure*}[t]
  \centering
  \includegraphics[width=\linewidth]{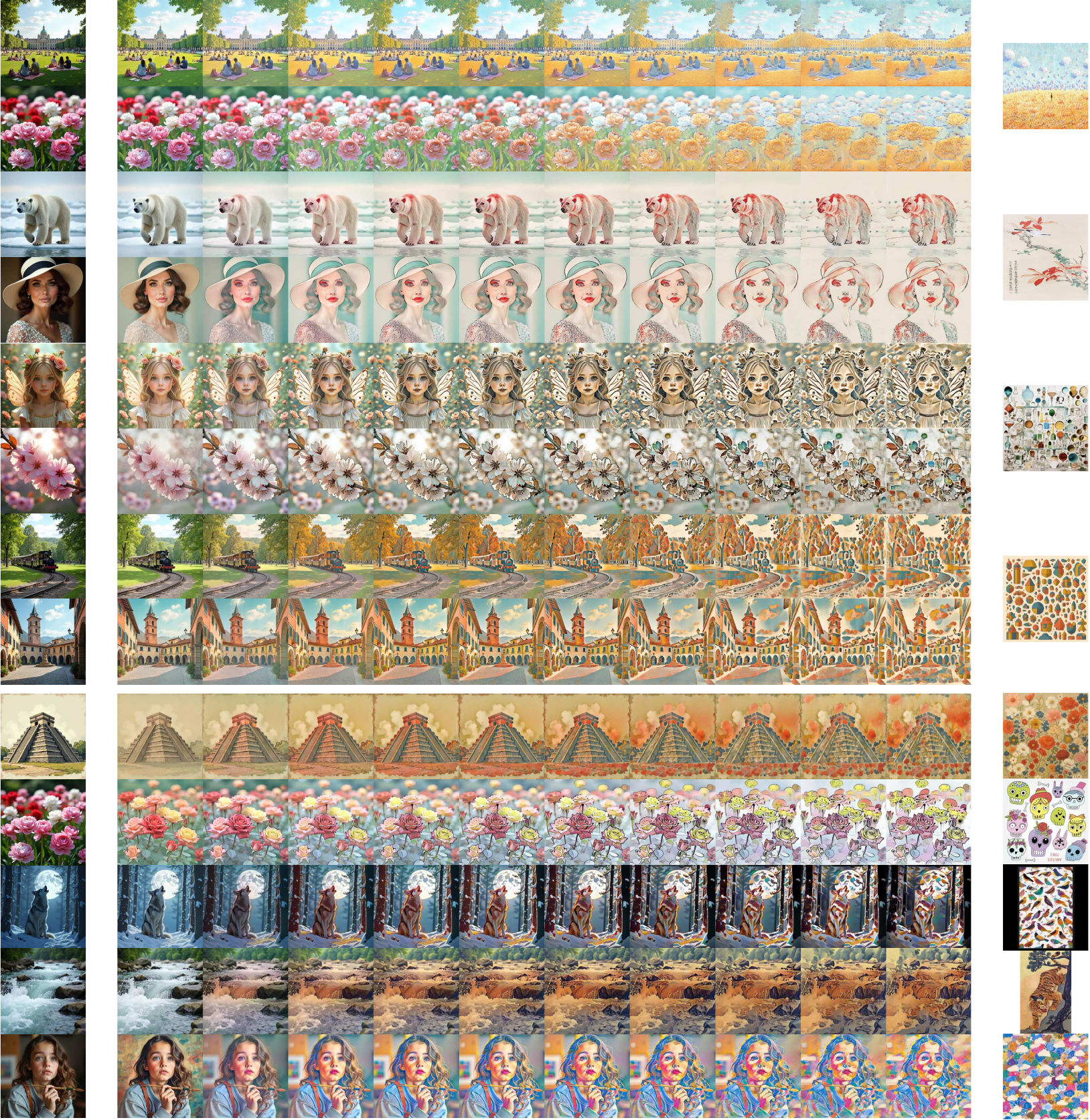}
  \caption{Additional continuous stylization results. Each row shows a content image on the left, outputs at monotonically increasing queried strengths in the middle, and the style reference on the right. Global appearance and local motifs are introduced progressively while the principal content structure remains stable.}
  \label{fig:style_ours}
\end{figure*}

This experiment tests whether projectors trained only at discrete anchors can generate visually ordered transitions at densely queried strengths.
Figure~\ref{fig:style_ours} shows content images on the left, style references on the right, and sequences produced by monotonically increasing the queried strength from left to right. The examples include people, animals, architecture, flowers, and natural scenes, with styles ranging from watercolor and ink drawing to object collages and repeated graphic motifs.

Across the sequences, changes occur progressively rather than as a single abrupt switch. Global color and tone typically appear first; contours, repeated elements, and denser textures become more prominent at larger strengths. Meanwhile, major structures---including the palace and pyramid outlines, the train track, facial identity, animal pose, and flower arrangement---remain aligned along each row. Even references with highly distinctive motifs, such as skulls or bottle-like objects, enter gradually instead of immediately replacing the content.

The ordered visual progression supports the intended role of spline interpolation: it extends a small set of learned anchor projectors to intermediate queries while maintaining a coherent content--style trajectory. Together with the user preference results, these examples show that the control variable is both interpretable and visually smooth.

\bibliography{main}
\end{document}